\documentclass[letterpaper, 10 pt, conference]{ieeeconf}
\IEEEoverridecommandlockouts
\usepackage{textcomp}
\usepackage{graphicx}
\usepackage{cite}
\usepackage[keeplastbox]{flushend} 
\usepackage{tikz}
\usepackage{calc}
\usepackage{amssymb}
\usepackage{pifont}
\newcommand{\xmark}{\ding{53}}%
\usepackage{diagbox} 
\usepackage{subcaption}

\title{\LARGE \bf Siamese Convolutional Neural Network for Sub-millimeter-accurate Camera Pose Estimation and Visual Servoing} \author{Cunjun Yu* \and
  Zhongang Cai* \and Hung Pham \and Quang-Cuong Pham
  \thanks{*Contributed equally} \thanks{The authors are with the
    School of Mechanical and Aerospace Engineering, Nanyang
    Technological University, Singapore.}  }

\begin{document}
\maketitle
\thispagestyle{empty}
\pagestyle{empty}

\begin{abstract}
  Visual Servoing (VS), where images taken from a camera typically
  attached to the robot end-effector are used to guide the robot
  motions, is an important technique to tackle robotic tasks that
  require a high level of accuracy. We propose a new neural network,
  based on a Siamese architecture, for highly accurate camera pose
  estimation. This, in turn, can be used as a final refinement step
  following a coarse VS or, if applied in an iterative manner, as a
  standalone VS on its own. The key feature of our neural network is
  that it outputs the relative pose between any pair of images, and
  does so with sub-millimeter accuracy. We show that our network can
  reduce pose estimation errors to 0.6 mm in translation and 0.4
  degrees in rotation, from initial errors of 10 mm / 5 degrees if
  applied once, or of several cm / tens of degrees if applied
  iteratively. The network can generalize to similar objects, is
  robust against changing lighting conditions, and to partial
  occlusions (when used iteratively). The high accuracy achieved
  enables tackling low-tolerance assembly tasks downstream: using our
  network, an industrial robot can achieve 97.5\% success rate on a
  VGA-connector insertion task without any force sensing
  mechanism. 

\end{abstract}

\section{Introduction}

Visual Servoing (VS), where images taken from a camera typically
attached to the robot end-effector are used to guide the robot
motions, is an important technique to tackle robotic tasks that
require a high level of accuracy~\cite{ref:tutorial}. Recently,
researchers have explored Deep Neural Networks (DNN) to implement
VS~\cite{ref:France, ref:endtoend}, with the hope that DNN can
mitigate certain shortcomings of classical VS, such as reliance on
manually-crafted features~\cite{ref:tutorial}, or sensitivity to
lighting conditions~\cite{ref:DVS}. The main issue with the
architecture proposed in~\cite{ref:France} is that a new network has
to be trained for each reference pose, which makes it unpractical for
actual industrial settings\footnote{In an insertion task, for
  instance, the relative pose between the camera and the object might
  change if the robot grasps objects in different ways . Under such
  circumstances, the reference pose has to be adjusted accordingly for
  a successful insertion.}.  Meanwhile, the approach proposed
in~\cite{ref:endtoend} has several-millimeter errors on synthetic data
and even larger errors on real data, which is insufficient for
low-tolerance industrial tasks.

Here we propose a new neural network for highly accurate camera pose
estimation, which can be used as a final refinement step following a
coarse VS or, if applied in an iterative manner, as a standalone VS on
its own. The key feature of our neural network is that it outputs the
relative pose between any pair of images (by contrast
with~\cite{ref:France}), and does so with sub-millimeter accuracy (by
contrast with~\cite{ref:endtoend}). This is achieved by leveraging a
\emph{Siamese} architecture~\cite{ref:siamese}, which
contains two branches of extractors (one per image). The features
extracted from the two images independently in this manner are then
compared at subsequent layers to achieve very high accuracy.

\begin{figure}[t]
\centering
  \includegraphics[width=\columnwidth]{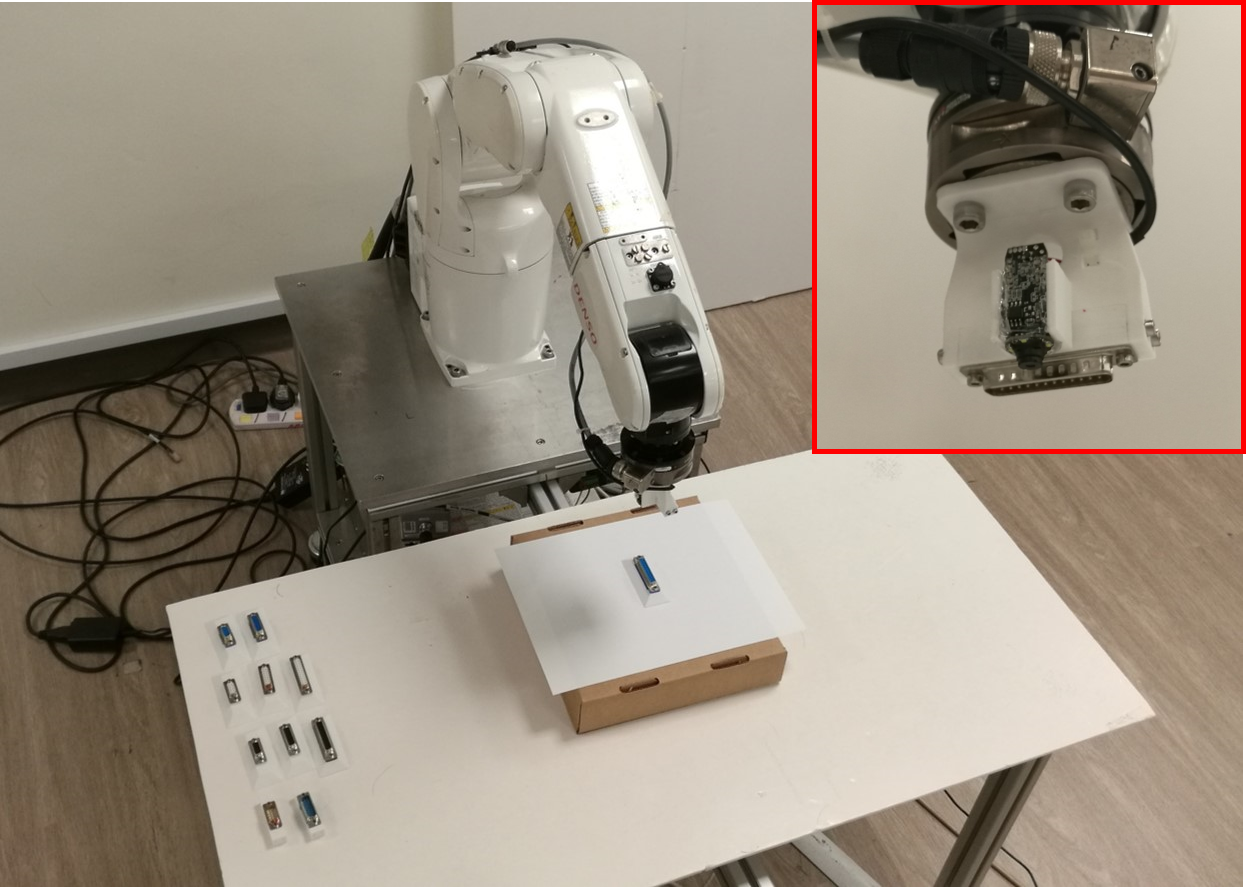}
\caption{An industrial robot is used for the insertion
  experiments. The male connector is attached to the end-effector with the
  camera. A better view of the end-effector is placed at the top right
  corner.}
\label{fig:set_up}
\end{figure}

More specifically, our main contributions are:
\begin{itemize}
\item our Siamese network can reduce initial errors of 10 mm in
  translation and 5 degrees in rotation to less than 0.6 mm in
  translation and 0.4 degrees in rotation, in a single shot;
\item even though the network is trained with small pose differences
  ($\leq$10 mm in translation $\leq$5 degrees in rotation), it can
  deal with large initial errors (several cms in translation and tens
  of degrees in rotation) when used in an iterative manner, as a
  standalone VS solution, without any compromise in final accuracy; 
  \item the high accuracy achieved enables tackling low-tolerance
  assembly tasks downstream: using our network, an industrial robot
  can achieve 97.5\% success rate on a VGA-connector insertion task
  without any force sensing mechanism;
\item our network can generalize to connectors it has never seen
  before, is robust against changing lighting conditions, and to
  partial occlusions (when used iteratively);
\end{itemize}

The paper is organized as follows. Section II discusses the related
work, in both classical and Deep Learning-based Visual
Servoing. Section III describes the network architecture in
detail. Section IV explains the method used to automatically collect
and accurately label the dataset used for training the model. Section
V reports the experiment results, on the test set and in a
physical VGA connector insertion task.

\section{Related Work}

\subsection{Camera pose estimation}

There has been a lot of research effort in camera pose estimation,
which is crucial for vision-based robotic manipulation. A popular
approach extracts feature points from two images, matches the
corresponding features and determines the relative camera pose
difference. Popular algorithms such as SIFT~\cite{ref:sift},
DAISY~\cite{ref:daisy}, SURF~\cite{ref:surf} and ORB~\cite{ref:orb}
are used in the local feature extraction and matching. However, the
accuracy of the classical approach may suffer due to a scarcity of
feature points found for matching.

In recent years, deep learning-based camera pose estimation has
attracted the attention of the research community. The accuracy of
deep learning-based camera pose estimation ranges from meters to
centimeters on different datasets. In~\cite{ref:posenet}, researchers
proposed a deep convolutional neural network with 2 m and
6\textdegree{} accuracy for large scale outdoor scenes and 0.5m and
10\textdegree{} for indoor scenes respectively. In
both~\cite{ref:relative_camera} and~\cite{ref:stiit}, the authors
trained and tested the neural networks proposed in previous works on
DTU dataset~\cite{ref:dataset} which consists of images pairs of
objects shot from different viewpoints, errors were reduced to a few
centimeters.

However, most of these works were designed for tasks like simultaneous
localization and mapping (SLAM) and visual odometry with relatively
large workspace domain; few papers have discussed utilizing the
robustness of deep learning for high precision camera pose estimation
which is fundamental for industrial applications such as a fine
assembly line.

\subsection{Deep learning-based Visual Servoing}

Deep learning-based visual servoing often performs camera pose
estimation iteratively while guiding the motion of the robot towards
the target pose so as to achieve high final accuracy.
In~\cite{ref:endtoend}, researchers generated a Visual Servoing Scene
Dataset (VSSD) by generating 5 simulated indoor scenes. Based on this
dataset, flownet~\cite{ref:flow} achieved very good performance,
reducing the error to a few millimeters.

In~\cite{ref:France}, an efficient way of generating dataset was
proposed to train robust neural networks for visual servoing was
proposed: by varying lighting conditions and adding random occlusions,
a dataset could be produced within hours. The neural network trained
on the dataset achieved sub-millimeter accuracy. However, the network
was trained to estimate the camera pose difference between the current
pose and a fixed reference pose by taking only one image captured at
the current base. This essentially means a new network has to be
trained for a new reference pose. In the same paper, another neural
network which takes in two images was proposed as an
extension. However, it could only achieve centimeter accuracy without
adopting photometric visual servoing~\cite{ref:photo} at the last
stage.

As pointed out by~\cite{ref:learnmove}, the awareness of egomotion
helps neural networks to learn features better. Therefore, in this
paper, we propose a neural network with Siamese architecture and takes
in two images from different viewpoints for high accuracy camera pose
estimation.

\section{Neural Network}

\subsection{Architecture}
\begin{figure*}[t]
\centering
  \includegraphics[width=\textwidth]{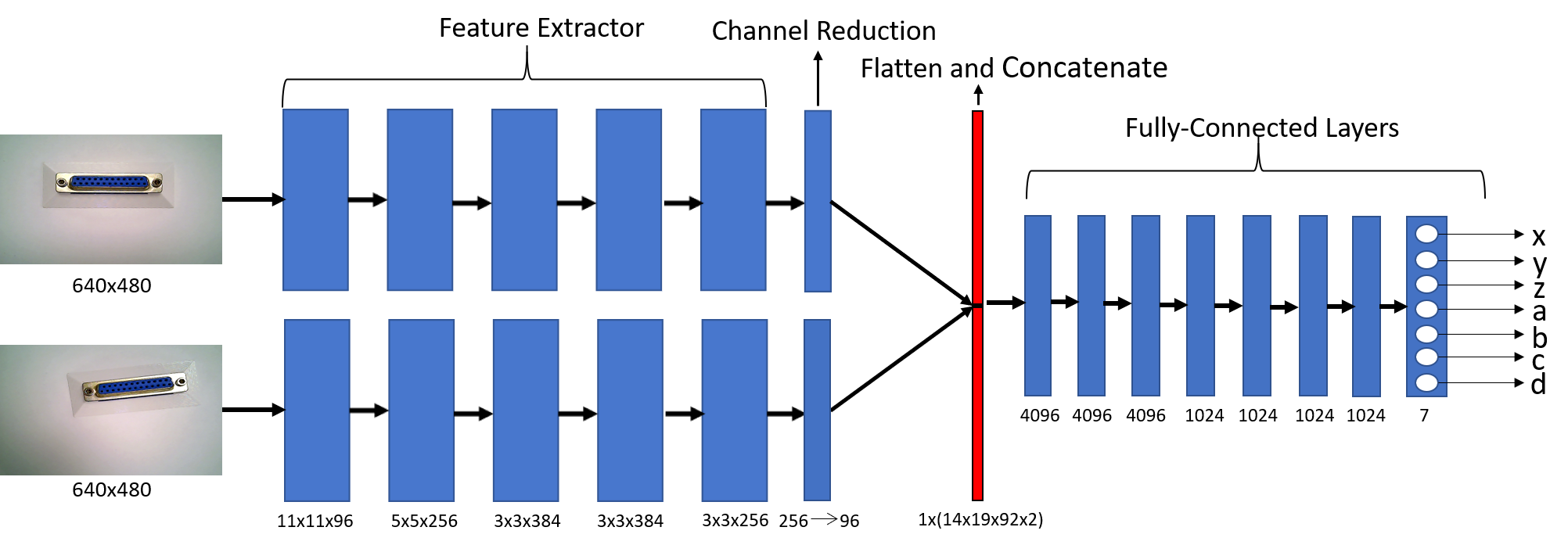}
\caption{Neural network architecture.}
\label{fig:verticalcell}
\end{figure*}

Since the network was designed to output the transformation between
two camera poses, it was intuitive to use a Siamese architecture (Fig
\ref{fig:verticalcell}) to extract features separately for the two
input images. In contrast to \cite{ref:flow}, which concatenates two images along the channel axis and performs feature extraction, applying convolutions on the two images is likely to produce independent features for matching at a later stage. Each arm adopts the feature extractor of CaffeNet (referred by many as AlexNet) for its simplicity and proven effectiveness on the \cite{ref:imagenet}.

To control the number of parameters, a channel reduction layer is used
to reduce channel from 256 to 96 with a convolutional kernel of size
1$\times$1.

We experimented many possible architecture designs for feature
matching. Direct operation of summation or subtraction of the two
feature maps from feature extractor has led to relatively poor
performances. Inspired by \cite{ref:goturn}, feature maps are
flattened, concatenated and fed into the classifier, which is adopted
from the Caffenet with minor adjustments to fit our input size. 5
additional fully connected layers are appended at the end to further
improve the performance.

For output layers, we attempted a two-branch-in and two-branch-out
architecture in order to estimate translations and rotations
separately. However, low accuracy was resulted. It can be explained
that the coupling effect of extrinsic parameters are significant and
hence, the network should output translation and rotation parameters
together, as in the final network design.

\subsection{Loss}

Since we designed the network to estimate the relative transformation $T_{\Delta}$ between \textit{any} two camera poses, not limiting to a fixed reference pose, an input pair is generated by collecting two samples, each consists of an image and a label $T_{d2e}$, which is the transformation from a default pose to the pose at which the image was taken (explained in greater details in the Dataset Generation and Training section). The images are marked as $I_{A}$ and $I_{B}$, the transformation label $T_{\Delta}$ is computed as:
$$T_{\Delta} = T_{d2e, A} \times T_{d2e, B}^{-1}$$

For brevity, the transformation label is decomposed into translation and rotation in terms of quaternions. The loss is then computed as:
$$L = \frac{1}{n}\sum_{i=1}^{n}wRMS(t_{i},{\hat{t}_{i}})+(1-w)RMS(q_{i},\hat{q}_{i})$$
where n is the number of input pairs, $t$ and $\hat{t}$ are translation label and estimation with the unit of meters, $q$ and $\hat{q}$ are rotation label and estimation in the form of normalized quaternions. Since the translation parameters are in meters, to balance the magnitude of values of translation and rotation parameters, a weight $w$ = 0.99 is used. $RMS$ is root-mean-square error defined as:
$$RMS(a, \hat{a}) = \sqrt{\frac{1}{m}\sum_{j=1}^{m}(a_{j}-\hat{a}_{j})^{2}}$$
where m is number of parameters. For translation, m = 3 whereas for rotation, m = 4 due to quaternion representation.

We chose quaternions over Euler angles because the variable space of roll, pitch, and yaw is not a good representation of the actual rotation; quaternions are a better way to encode the axis-angle representation. This is also supported by experiments in some previous works such as \cite{ref:regnet}.

\section{Set-up}

As shown in Figure \ref{fig:set_up}, we designed and 3D printed
an end-effector to be mounted at the tip of the robotic arm. A male
connector can be mounted on the bottom of the end-effector. The
corresponding female connector is mounted on a white frustum-shape
base that is placed on a white A4-sized paper. The base is required as
some connectors cannot stand on its own.

Clearly the base can provide some features for the network, however,
we argue that the base can be regarded as part of the connector. In
the actual production environment, there will always be additional
features (such as other machine parts) in the scene other than the
object of interest. We chose white as the color for the base and the
background for two reasons: firstly, noise can be added easily to
white pixels for data augmentation. Secondly, textureless background
and base provide fewer features for the neural network to learn from
so as to encourage the network to focus on the connectors.

An inexpensive (less than \$20) short-range camera with the field of
view 70 \textdegree and the resolution 640$\times$480, was installed
on the side of the end-effector. The camera was equipped with 4 LED
lights to improve lighting as the base was usually in the shadow of the
robotic arm and the end-effector. We did not select an expensive
high-resolution camera for three reasons: the model should not rely on
high-resolution to reduce the cost of future industrial application;
the neural network was designed to be light-weight so a small input
size was preferred; the camera had to be small so it could be placed
near the male connector such that the female connector could appear in
its view at the insertion pose.

\section{Dataset Generation and Training}
\subsection{Sample collection}
\begin{figure}[ht]
\centering
  \includegraphics[width=\columnwidth]{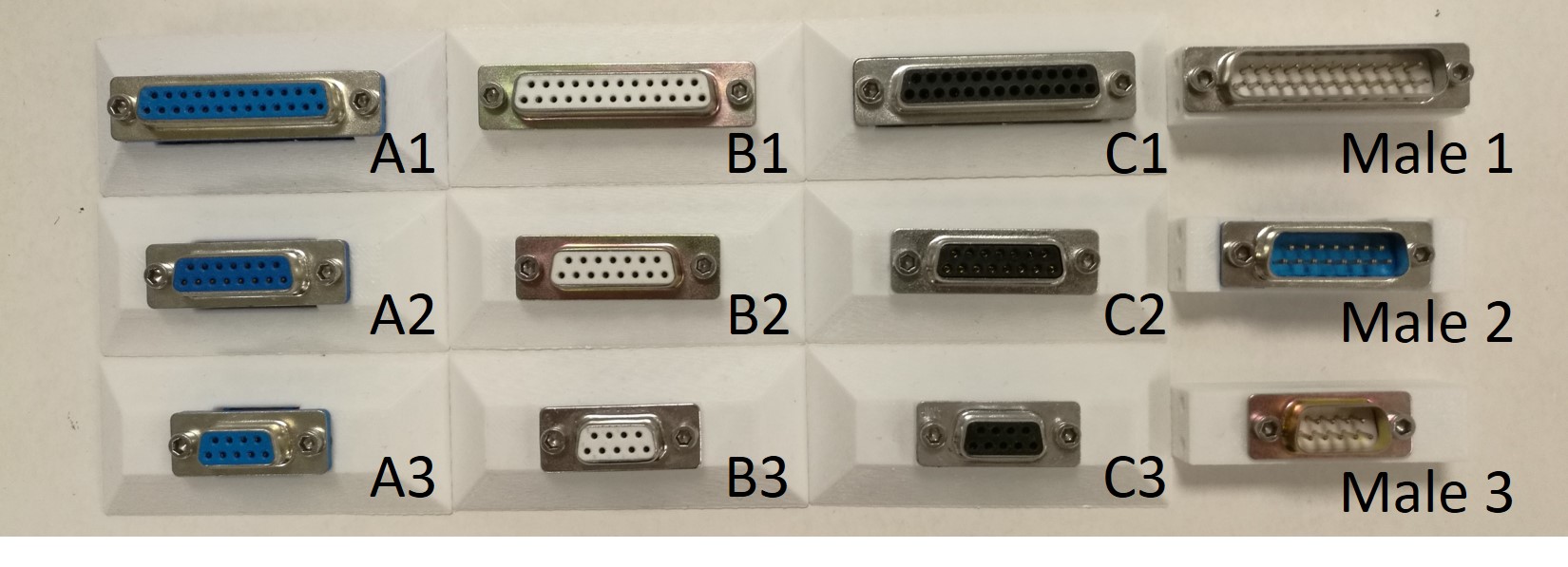}
  \caption{The female connectors come  with different shapes (A: blue;
    B: white; C: black)  and sizes (1,2 and 3); the  male connectors of
    corresponding sizes are used.}
\label{fig:connectors}
\end{figure}

Figure \ref{fig:connectors} shows the connectors used in the dataset and the experiments. 

We guided the robotic arm to an insertion pose where the male
connector mounted on the end-effector was inserted into the female
connector on the base. Then the end-effector was lifted vertically for
15 cm to allow the camera to capture the full view of the connector
with sufficient margins. The pose of the end-effector can be found by
forward kinematics and was recorded as the default pose $T_{0}$. The
default pose had its z-axis pointing vertically downwards if see from
the world frame. Note that the default pose is not the reference pose
mentioned in the experiments.

The entire collection process was automated and samples were generated
by randomly changing the end-effector pose around the default
pose. Since we were focusing on the \textit{last step} of visual
servoing, the sampling range of translation and rotation were small:
the origin of new camera pose was uniformly sampled within a vertical
cylinder with radius 5 mm and height 10 mm, with the default pose
frame origin at the center of the bottom of the cylinder. The rotation
was sampled uniformly from -5 degrees to 5 degrees for roll and
pitch and -10 degrees to 10 degrees for yaw. An image taken by the
camera and a corresponding pose (in the form of a transformation
matrix from the default pose to the new end-effector pose $T_{d2e}$)
were recorded for each arbitrary pose.

\subsection{Datasets}

We firstly collected a small dataset with only the connector type
A1 to verify the network's robustness against changes in the lighting condition. Due to the shadows of the robotic arm and the end-effector, lighting was vastly different if the female connector is moved into the shadow. Rotation was also needed such that the shadows do not always appear at the same corner of a reference image. In addition, the reflection pattern on the female connectors were strongly affected by the position and the direction of the connector. In this dataset, we placed the female connector at 5 points on the workbench, with four spaced 5 cm in both x and y directions from each other forming the vertices of a square of, and the last one at the center of the square. On each point, we rotated the base 0\textdegree, 30\textdegree, 60\textdegree and 90\textdegree and for each orientation, 200 samples were collected, totalling $5\times4\times200=4000$ samples. 

In addition, to enable generalization across different connectors with a single network, 1000 samples were collected for each of the 9 female connectors (A1, A2, ... ,C3) to form a larger dataset. Further details are covered below.

\subsection{Training}
\label{sec:training}

For each connector, we took 50 samples each for validation and testing respectively, 
the rest was used for training. Note that for $n$
samples in each set, $n^{2}$ input pairs can be created. To prevent
taking too much disk space, the input pair were created during
training instead of being pre-processed.

We trained two models $M_{A1}$ and $M_8$. Model $M_{A1}$ was trained with
only A1 ($M_{A1}$) for 10 epochs. A quarter of
the maximum number of the training set, $3900^{2}\times0.25=3802500$
input pairs were used in the training. The learning rate was $10^{-4}$
initially, and halved after the $4^{th}$, $6^{th}$ and $8^{th}$
epoch. Adam optimizer was used with $\beta_{1} = 0.9$,
$\beta_{2} = 0.999$, $\epsilon = 10^{-8}$ and no weight decay. The
training was run on 4 GTX-1080Ti with a batch size of 256. Uniform
weights initialization was applied.

Model $M_8$ was trained to evaluate the network's ability to generalize
across different connectors. We selected 8 connectors (all except A2)
to form a dataset of $8\times900^{2}=6480000$ input pairs for training. 
A2 was used in the experiment to verify the robustness of the network against
a novel object.

\section{Experiment Results}
We evaluated the performance of models $M_{A1}$ and $M_{8}$ first on the test sets
(as described in Sec.~\ref{sec:training}) to
obtain the mean errors of each parameter for a qualitative analysis of
the model. After that, we tested the models on actual insertion tasks.

\begin{figure}[ht]
\centering
  \includegraphics[width=\columnwidth]{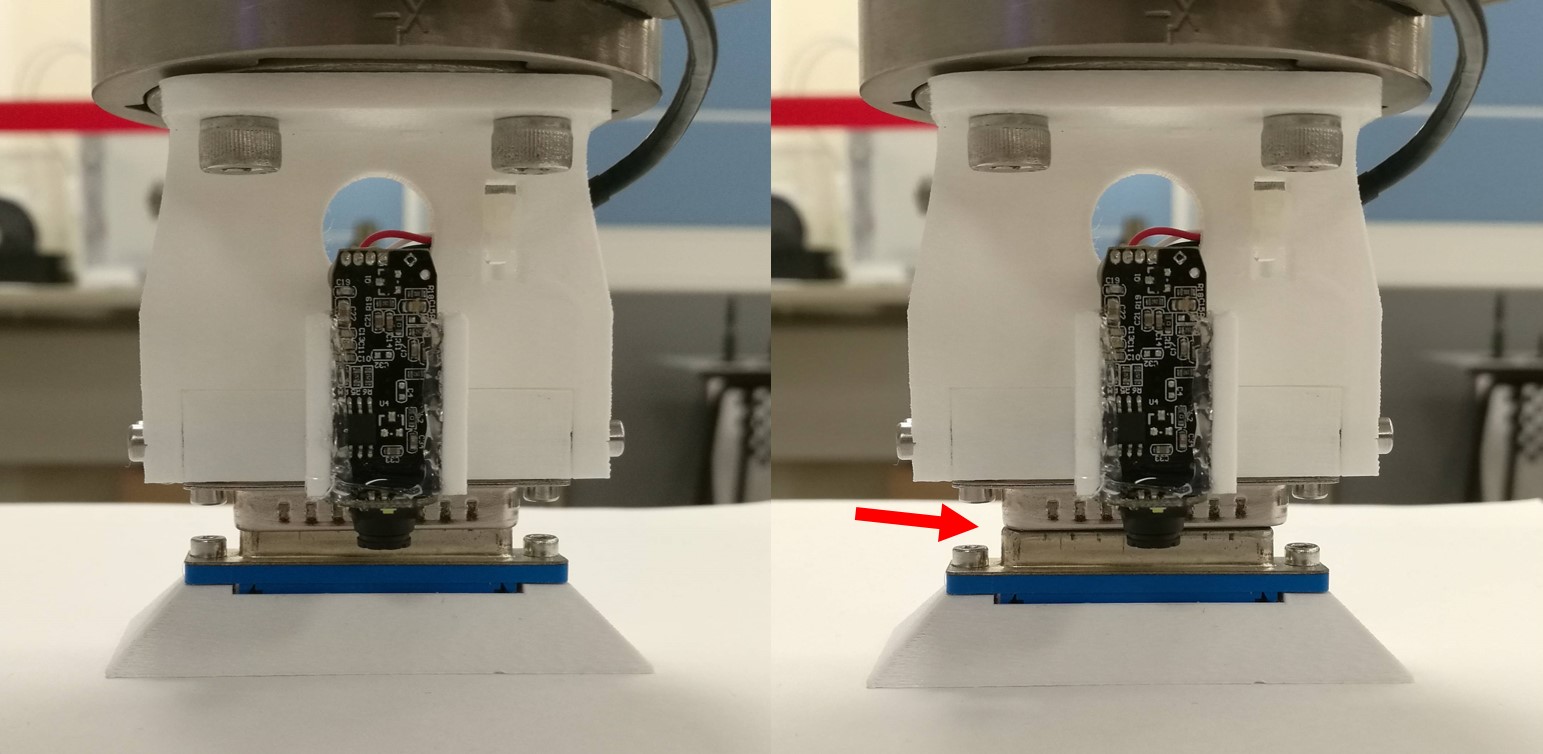}
\caption{A successful (left) and an unsuccessful (right, collision point indicated by the red arrow) insertion.}
\label{fig:success_fail}
\end{figure}

As shown in Figure \ref{fig:success_fail}, an
insertion is successful only if the male connector could completely go into the female
counterpart. To evaluate the range of tolerances that allow successful
insertions, we manually added
offsets to the reference pose (the pose that results in a successful
insertion) and tested the insertion. However, evaluating parameters separately (one at a time)
is meaningless due to their coupled effect in the actual
insertion. Hence, we clustered translation parameters (x, y, and z)
and rotation parameters (roll, pitch, and yaw) separately into two
groups. Each time we varied all parameters in a group by the same
amount. To reduce the number of combinations, all offsets tested were
positive. Table \ref{tab:thresholds} was produced using connector
A1. Note that this experiment was meant to allow readers to have a
rough idea about the difficulty of the insertion; the coupled effect
is much more complicated in the actual task.

It was observed that insertion of VGA connectors is indeed a
challenging task: to achieve a successful insertion, a large error in
certain parameters must be compensated by an extremely small error in
other parameters.

\begin{table}[ht]
  \caption{Insertion tolerance analysis: a qualitative evaluation of the difficulty of the task. Note that these are not the insertion experiment results. Translation (x, y and z) and rotation (roll, pitch and yaw) parameters were clustered into two groups and the same offset was applied on parameters in the same group each time; A tick/cross indicates at that translation and rotation offsets, the insertion was successful/unsuccessful. For example, if x, y and z are all offset by +0.3 mm whereas roll, pitch and yaw are all offset by +1.00 degree, the insertion will be unsuccessful.}

\begin{center}
\begin{tabular}{|c|c|c|c|c|c|c|c|c|}
\hline
\textbf{\backslashbox{(deg)}{(mm)}}&\textbf{0.0}&\textbf{0.1}&\textbf{0.2}&\textbf{0.3}&\textbf{0.4}&\textbf{0.5}&\textbf{0.6}&\textbf{0.7}  \\
\hline
\textbf{0.00}&\checkmark&\checkmark&\checkmark&\checkmark&\checkmark&\checkmark&\checkmark&\xmark \\
\hline
\textbf{0.25}&\checkmark&\checkmark&\checkmark&\checkmark&\checkmark&\checkmark&\checkmark&\xmark \\
\hline
\textbf{0.50}&\checkmark&\checkmark&\checkmark&\checkmark&\checkmark&\xmark&\xmark&\xmark\\
\hline 
\textbf{0.75}&\checkmark&\checkmark&\checkmark&\checkmark&\xmark&\xmark&\xmark&\xmark\\
\hline
\textbf{1.00}&\checkmark&\checkmark&\checkmark&\xmark&\xmark&\xmark&\xmark&\xmark\\
\hline
\textbf{1.25}&\checkmark&\checkmark&\checkmark&\xmark&\xmark&\xmark&\xmark&\xmark\\
\hline
\textbf{1.50}&\checkmark&\checkmark&\xmark&\xmark&\xmark&\xmark&\xmark&\xmark\\
\hline
\textbf{1.75}&\checkmark&\checkmark&\xmark&\xmark&\xmark&\xmark&\xmark&\xmark\\  
\hline
\textbf{2.00}&\checkmark&\checkmark&\xmark&\xmark&\xmark&\xmark&\xmark&\xmark\\
\hline
\textbf{2.25}&\xmark&\xmark&\xmark&\xmark&\xmark&\xmark&\xmark&\xmark \\
\hline
\end{tabular}
\label{tab:thresholds}
\end{center}
\end{table}

\subsection{Model trained on A1 ($M_{A1}$) for the evaluation of robustness against changes in the lighting conditions}
\label{sec:modelA1}
\subsubsection{Performance on the test set}
the test set of A1 consists of $50^{2}=2500$ input pairs. The mean
absolute errors of translation and rotation across the entire test set
are tabulated in Table \ref{tab:A1_testing_set_perf}. For easy
interpretation, the quaternions were first converted to Euler angles
for rotation error computation.

\begin{table}[ht]
\caption{Errors of model $M_{A1}$ on the test set}
\begin{center}
\begin{tabular}{|c|c|c|c|c|c|c|}
\hline
\textbf{Object}&$\mathbf{e_{x}}$\textbf{/mm}&$\mathbf{e_{y}}$\textbf{/mm}&$\mathbf{e_{z}}$\textbf{/mm}&$\mathbf{e_{\phi}}$\textbf{/\textdegree}&$\mathbf{e_{\theta}}$\textbf{/\textdegree}&$\mathbf{e_{\psi}}$\textbf{/\textdegree} \\
\hline 
A1&0.34&0.36&0.44&0.19&0.18&0.19\\
\hline
\end{tabular}
\label{tab:A1_testing_set_perf}
\end{center}
\end{table}

Notice that the mean translation errors were below 0.5 mm; the mean
rotation errors were smaller than 0.2 degree. In the experiments,
rotation affected the view of the camera more than translation did,
hence there was no surprise that the network could learn to estimate
the rotation extremely well. The errors in translation were more
subtle, yet the network could still give reasonably good translation
estimations.

\subsubsection{Performance on actual insertion}
For the actual insertion experiment, we firstly adjusted the robotic
arm and the base such that if the end-effector went vertically
downwards, the male connector could fully go into the female
connector. An image $I_{ref}$ was then captured. Note that $I_{ref}$
was taken only once at the start of the experiment.

We then manually shifted (with rotation) the female connector on the
workbench slightly and let a second image $I_{test}$ to be taken. The
network took in $I_{ref}$ as $I_{A}$ and $I_{test}$ as $I_{B}$ and
output 7 parameters to construct a transformation matrix $T_{\Delta}$.

Afterwards, the robot was moved to $T_{est}$ which is obtained as:
$$T_{est} = T_{\Delta} \times T_{test}$$
If the estimation was perfect, the connector in the third image
$I_{est}$ taken would look identical as the connector in $I_{ref}$.

The end-effector is then descended \textit{blindly} to attempt the
insertion: no any other intermediate adjustments or corrections were
performed. To expedite experiment, we did not apply further force to
achieve a firm connection between the two connectors. Once the male
connector fully went into the female connector, it was counted as a
successful attempt. If the male connector got stuck at the rim of the
female connector, it was counted as unsuccessful.

We repeated the above steps 50 times and the results are tabulated in
\ref{tab:A1_actual_insert_perf}. Due to the displacement of the female
connector, $I_{test}$ could be different from $I_{ref}$ in terms of
lighting conditions due to shadows and the difference between
$I_{ref}$ and $I_{test}$ could become larger and larger since
$I_{ref}$ was constant. Yet, the network was able to catch up with the
movement and performed most of the insertion correctly.

\begin{table}[ht]
\caption{Performance of model $M_{A1}$ on actual insertion}
\begin{center}
\begin{tabular}{|c|c|c|c|}
\hline
 &\textbf{\#Success}&\textbf{\#Total Attempts}&\textbf{Percentage(\%)}\\
\hline 
A1&47&50&94\\
\hline
\end{tabular}
\label{tab:A1_actual_insert_perf}
\end{center}
\end{table}

\subsection{Model trained on 8 connectors ($M_{8}$) for the evaluation of robustness against a novel connector}
\label{sec:modelM8}
\subsubsection{Performance on the test set}
\begin{figure*}[ht]
  \begin{subfigure}{\columnwidth}
    \includegraphics[width=\linewidth]{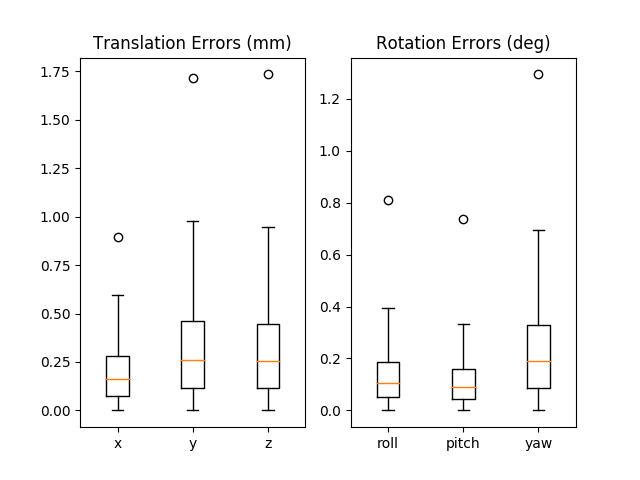}
    \caption{Distributions of translation errors and rotation errors
      represented by box plots for the 8 connectors. The maximum
      outliers are drawn as circles on the top. The five horizontal
      lines from top to bottom indicate the maximum fence, the third
      quartile, the mean value, the first quartile and the minimum
      fence.}
  \end{subfigure}
  \hspace{\fill}  
  \begin{subfigure}{\columnwidth}
    \includegraphics[width=\linewidth]{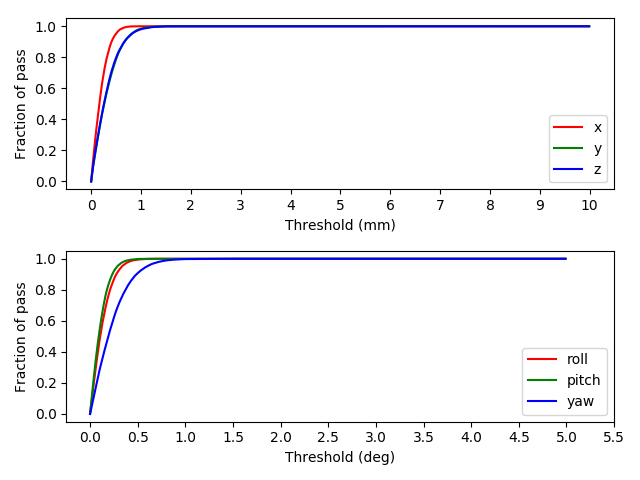}
    \caption{Percentage of pass vs threshold curves for the 8
      connectors. Top: translation errors; bottom: rotation
      errors. The graph shows the percentage of test pair with error
      lower that the corresponding threshold values.}
  \end{subfigure}
  \begin{subfigure}{\columnwidth}
    \includegraphics[width=\linewidth]{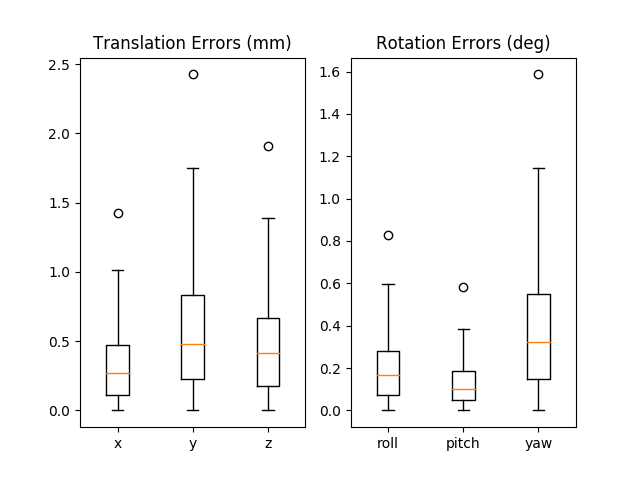}
    \caption{Distributions of translation errors and rotation errors
      represented by box plots for connector A2. Note that A2 is not
      involved in the training set.}
  \end{subfigure}
  \hspace{\fill}  
  \begin{subfigure}{\columnwidth}
    \includegraphics[width=\linewidth]{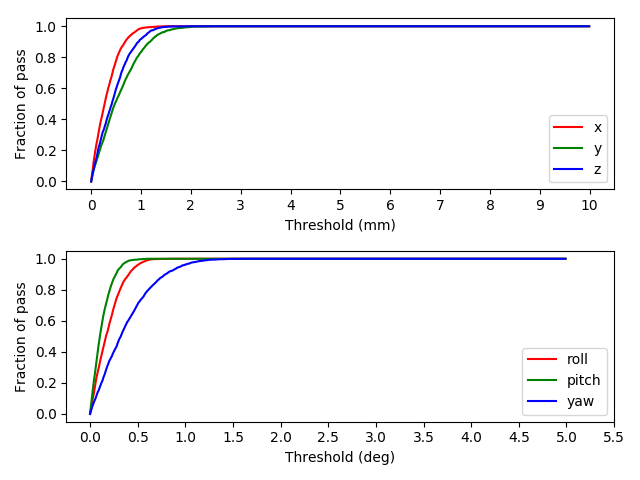}
    \caption{Percentage of pass vs threshold curves for connector
      A2. Note that A2 is not involved in the training set.}
  \end{subfigure}
  \caption{Quantitative experiment results of $M_{8}$ on the test set}
  \label{fig:test_set_perf}
\end{figure*}

Similar to experiments conducted on $M_{A1}$, we fed the trained model
$M_{A1}$ with the test set which consists of $8\times50^{2}=20000$
input pairs. Results are tabulated in Table
\ref{tab:8_testing_set_perf} and Figure \ref{fig:test_set_perf}.

\begin{table}[ht]
\caption{Errors of model $M_{8}$ on the test set}
\begin{center}
\begin{tabular}{|c|c|c|c|c|c|c|}
\hline
\textbf{Object}&$\mathbf{e_{x}}$\textbf{/mm}&$\mathbf{e_{y}}$\textbf{/mm}&$\mathbf{e_{z}}$\textbf{/mm}&$\mathbf{e_{\phi}}$\textbf{/\textdegree}&$\mathbf{e_{\theta}}$\textbf{/\textdegree}&$\mathbf{e_{\psi}}$\textbf{/\textdegree} \\
\hline 
A1&0.17&0.30&0.23&0.12&0.13&0.26 \\
\hline
\textbf{A2}&\textbf{0.36}&\textbf{0.55}&\textbf{0.43}&\textbf{0.21}&\textbf{0.14}&\textbf{0.42}  \\ 
\hline
A3&0.23&0.29&0.32&0.13&0.12&0.25 \\
\hline
B1&0.19&0.32&0.25&0.14&0.11&0.25 \\
\hline 
B2&0.22&0.34&0.36&0.15&0.12&0.25 \\
\hline
B3&0.18&0.39&0.40&0.13&0.11&0.21  \\ 
\hline
C1&0.20&0.34&0.24&0.14&0.11&0.21 \\
\hline
C2&0.16&0.28&0.33&0.11&0.10&0.22 \\
\hline
C3&0.20&0.28&0.37&0.11&0.10&0.21 \\
\hline
\end{tabular}
\label{tab:8_testing_set_perf}
\end{center}
\end{table}

In Figure \ref{fig:test_set_perf}, the model has achieved very high
accuracy on the 8 connectors that the network has seen in the
training: the translation estimations were around 0.25 mm or better on
average, and rotation estimations were around 0.2 degrees or better on
average. Except for very few outliers, the vast majority of errors
fell below 1.0 mm and 0.8 degree resulting in a steep fraction of pass
versus threshold graphs.

The performance on the test set of A2, which was novel to the network,
was not as good as that of the other 8 connectors but still highly
accurate. The mean errors reached around 0.5 mm and 0.3 degrees. The
fraction of pass versus threshold graphs are not as steep but the vast
majority of errors still fall below 1.7 mm and 1.2 degrees.

\subsubsection{Performance on actual insertion}

similar to the experiments conducted with the model $M_{A1}$, we
firstly adjusted the pose the robot to obtain $I_{ref}$. We also
defined this end-effector pose as the reference pose, $T_{ref}$.

Instead of moving base like in the previous experiment which required
human intervention, we automated this experiment by randomly moving
the end-effector.

Note that the network can estimate the camera displacement
$T_{\Delta}$ from any initial pose A to current pose B, but in order
to perform a successful insertion, the reference pose $T_{ref}$ was
always used as the pose A.

We then randomly choose a new pose the same way in the Sample
Collection and Dataset section, we called it the test pose $T_{test}$.
An image $I_{test}$ was captured at the test pose. The network was
tasked to estimate a transformation $T_{\Delta}$ to move the pose back
to the reference pose.

$T_{est}$ was computed the same way as that in the experiment with
$M_{A1}$. If the estimation was perfect, $T_{est}$ should coincide
with $T_{ref}$.

The end-effector was then moved to $T_{est}$ followed by descending
end-effector to attempt the insertion. For each connector, 25 trials
were conducted running on model $M_{8}$.
\begin{table}[ht]
\caption{Performance of model $M_{8}$ on actual insertion}
\begin{center}
\begin{tabular}{|c|c|c|c|}
\hline
 &\textbf{\#Success}&\textbf{\#Total Attempts}&\textbf{Percentage(\%)}\\
\hline 
A1&24&25&96\\
\hline
\textbf{A2}&\textbf{24}&\textbf{25}&\textbf{96}\\ 
\hline
A3&24&25&96 \\
\hline
B1&25&25&100 \\
\hline 
B2&23&25&92 \\
\hline
B3&25&25&100 \\ 
\hline
C1&25&25&100 \\
\hline
C2&25&25&100 \\
\hline
C3&25&25&100 \\
\hline
Overall&195&200&97.5 \\
\hline
\end{tabular}
\label{tab:8_actual_insert_perf}
\end{center}
\end{table} 

Table \ref{tab:8_actual_insert_perf} shows that the network was able
to achieve extremely high accuracy. Note that the connector A2 (in
bold) was not involved in the train set, yet, the network was
performing very well on this novel connector after learning similar
connectors.

\subsection{Iterative estimation for actual insertion with model $M_{A1}$}

\begin{figure*}[t]
\centering
  \includegraphics[width=\textwidth]{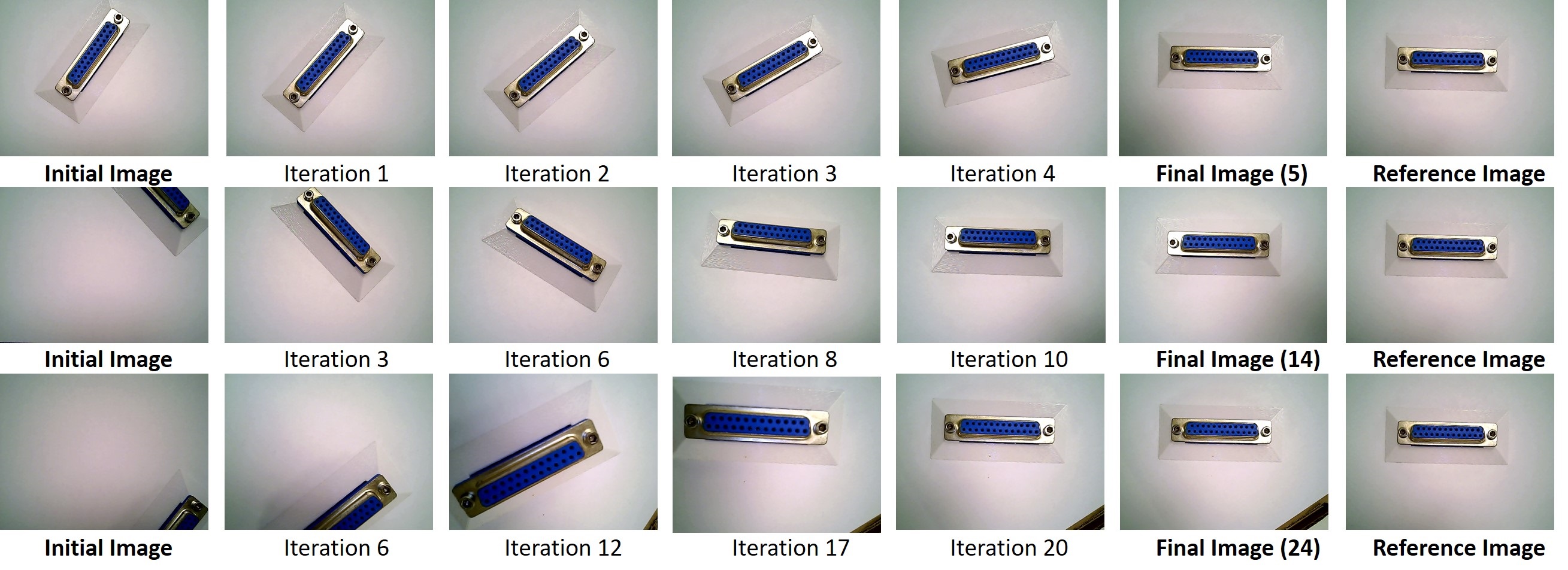}
  \caption{By performing iterative estimations, model $M_{A1}$ was able to guide
  the robot from extreme starting poses to the reference pose, achieving successful insertions. Selected images from an easy, a medium and a hard test case are shown in the first, the second and the third row respectively (see Table \ref{tab:iter_perf} for details). Note that model $M_{A1}$ was not trained to estimate pose difference this large. The reference image (in the right-most column) was fixed in the experiment. The number in the bracket of each Final Image caption is the total number of iterations needed to reach the final convergence.}
\label{fig:iter_perf}
\end{figure*}

In the insertion experiments reported in Sec~\ref{sec:modelA1} and~\ref{sec:modelM8}, the differences 
between the initial poses and the reference
poses of the robot fell within the sampling range of the train set. 
Hence, the models' one-shot estimations were sufficiently precise to 
achieve successful insertions.

In this third experiment, however, we used model $M_{A1}$ to estimate
pose differences that were much larger than the sampling range in the 
train set. We purposely placed the female connector
far away from the image center, such that the pose difference 
were much larger than transformation labels in the train
set. Furthermore, in two of the three test cases, the connectors were even partially
occluded in the initial camera images.

Although model $M_{A1}$ could not provide accurate one-shot estimations, it was 
\textit{always} able to move the robot closer to the
reference pose. Eventually through iterations, the reference pose was
\textit{always} reached in all test cases, allowing successful insertions.
The results are tabulated in \ref{tab:iter_perf}. Camera images at selected iterations are
shown in Figure \ref{fig:iter_perf}.

\begin{table}[ht]
\caption{Iterative estimation performance. Number of iterations
required to converge with various percentage of the female connector
visible in the first image. All insertions were successful.}
\begin{center}
\begin{tabular}{|c|c|c|c|c|} \hline
\textbf{Difficulty}&\textbf{Percentage visible}&\textbf{Test
1}&\textbf{Test 2}&\textbf{Test 3}\\ \hline
Easy&100\%&5&4&8 \\
\hline 
Medium&50\%&14&17&16 \\
\hline 
Hard&30\%&18&24&15 \\
\hline
\end{tabular}
\label{tab:iter_perf}
\end{center}
\end{table}

Moreover, Figure \ref{fig:iter_perf} shows that the network is very
robust against changing lighting conditions as the final image of the
second row has different reflection pattern compared to the reference
image. The network is also robust against noises such as blurry image
(iteration 17 of the last row was captured when the camera was very close to the
connector) and minor changes in the background (iteration 12-24 of the
last row, the edge of the A4-sized paper was captured and this was
never included in the training set).

\section{Conclusion} 

We present a Siamese convolutional neural network
as the last piece to complete the jigsaw of deep learning-based visual
servoing. By one-shot estimation, it achieves extremely high accuracy
in camera pose estimation (less than 0.6 mm in translation and 0.4 degrees in rotation), which makes it a plausible solution to difficult
real-world application such as low-tolerance insertion.

In addition, the network is able to handle large pose difference if used iteratively, even it has only been trained to handle the fine difference. This makes the network a standalone visual servoing solution.

Instead of running evaluations on the test sets, we have demonstrated the model's exceptional performance (97.5\% success rate) on actual insertion task with VGA-connectors. The model is even robust against changing light conditions and able to handle a novel connector through training of a few similar counterparts.

In the future, various scenes can be furthered added to the dataset by
replacing the white pixels with other colors to improve generalization
across different environments.


\bibliographystyle{IEEEtran}
\bibliography{references}

\begin{thebibliography}{10}
\providecommand{\url}[1]{#1}
\csname url@samestyle\endcsname
\providecommand{\newblock}{\relax}
\providecommand{\bibinfo}[2]{#2}
\providecommand{\BIBentrySTDinterwordspacing}{\spaceskip=0pt\relax}
\providecommand{\BIBentryALTinterwordstretchfactor}{4}
\providecommand{\BIBentryALTinterwordspacing}{\spaceskip=\fontdimen2\font plus
\BIBentryALTinterwordstretchfactor\fontdimen3\font minus
  \fontdimen4\font\relax}
\providecommand{\BIBforeignlanguage}[2]{{%
\expandafter\ifx\csname l@#1\endcsname\relax
\typeout{** WARNING: IEEEtran.bst: No hyphenation pattern has been}%
\typeout{** loaded for the language `#1'. Using the pattern for}%
\typeout{** the default language instead.}%
\else
\language=\csname l@#1\endcsname
\fi
#2}}
\providecommand{\BIBdecl}{\relax}
\BIBdecl

\bibitem{ref:tutorial}
S.~{Hutchinson}, G.~D. {Hager}, and P.~I. {Corke}, ``A tutorial on visual servo
  control,'' \emph{IEEE Transactions on Robotics and Automation}, vol.~12,
  no.~5, pp. 651--670, Oct 1996.

\bibitem{ref:France}
Q.~{Bateux}, E.~{Marchand}, J.~{Leitner}, F.~{Chaumette}, and P.~{Corke},
  ``Training deep neural networks for visual servoing,'' in \emph{2018 IEEE
  International Conference on Robotics and Automation (ICRA)}, May 2018, pp.
  1--8.

\bibitem{ref:endtoend}
A.~{Saxena}, H.~{Pandya}, G.~{Kumar}, A.~{Gaud}, and K.~M. {Krishna},
  ``Exploring convolutional networks for end-to-end visual servoing,'' in
  \emph{2017 IEEE International Conference on Robotics and Automation (ICRA)},
  May 2017, pp. 3817--3823.

\bibitem{ref:DVS}
K.~Deguchi, ``A direct interpretation of dynamic images with camera and object
  motions for vision guided robot control,'' \emph{International Journal of
  Computer Vision}, vol.~37, pp. 7--20, 06 2000.

\bibitem{ref:siamese}
J.~Bromley, I.~Guyon, Y.~LeCun, E.~S\"{a}ckinger, and R.~Shah, ``Signature
  verification using a "siamese" time delay neural network,'' in
  \emph{Proceedings of the 6th International Conference on Neural Information
  Processing Systems}, ser. NIPS'93.\hskip 1em plus 0.5em minus 0.4em\relax San
  Francisco, CA, USA: Morgan Kaufmann Publishers Inc., 1993, pp. 737--744.

\bibitem{ref:sift}
\BIBentryALTinterwordspacing
D.~G. Lowe, ``Distinctive image features from scale-invariant keypoints,''
  \emph{Int. J. Comput. Vision}, vol.~60, no.~2, pp. 91--110, Nov. 2004.
  [Online]. Available: \url{https://doi.org/10.1023/B:VISI.0000029664.99615.94}
\BIBentrySTDinterwordspacing

\bibitem{ref:daisy}
E.~{Tola}, V.~{Lepetit}, and P.~{Fua}, ``Daisy: An efficient dense descriptor
  applied to wide-baseline stereo,'' \emph{IEEE Transactions on Pattern
  Analysis and Machine Intelligence}, vol.~32, no.~5, pp. 815--830, May 2010.

\bibitem{ref:surf}
\BIBentryALTinterwordspacing
H.~Bay, A.~Ess, T.~Tuytelaars, and L.~Van~Gool, ``Speeded-up robust features
  (surf),'' \emph{Comput. Vis. Image Underst.}, vol. 110, no.~3, pp. 346--359,
  Jun. 2008. [Online]. Available:
  \url{http://dx.doi.org/10.1016/j.cviu.2007.09.014}
\BIBentrySTDinterwordspacing

\bibitem{ref:orb}
E.~{Rublee}, V.~{Rabaud}, K.~{Konolige}, and G.~{Bradski}, ``Orb: An efficient
  alternative to sift or surf,'' in \emph{2011 International Conference on
  Computer Vision}, Nov 2011, pp. 2564--2571.

\bibitem{ref:posenet}
A.~{Kendall}, M.~{Grimes}, and R.~{Cipolla}, ``Posenet: A convolutional network
  for real-time 6-dof camera relocalization,'' in \emph{2015 IEEE International
  Conference on Computer Vision (ICCV)}, Dec 2015, pp. 2938--2946.

\bibitem{ref:relative_camera}
I.~Melekhov, J.~Ylioinas, J.~Kannala, and E.~Rahtu,
  ``\BIBforeignlanguage{English}{Relative camera pose estimation using
  convolutional neural networks},'' in
  \emph{\BIBforeignlanguage{English}{Advanced Concepts for Intelligent Vision
  Systems - 18th International Conference, ACIVS 2017, Proceedings}}, ser.
  Lecture Notes in Computer Science.\hskip 1em plus 0.5em minus 0.4em\relax
  Germany: Springer Verlag, 2017, pp. 675--687, jufoid=62555.

\bibitem{ref:stiit}
L.~J. Charco, B.~X. Vintimilla, and A.~D. Sappa, ``Deep learning based camera
  pose estimation in multi-view environment,'' in \emph{14th IEEE International
  Conference on Signal Image Technology \& Internet based Systems (SITIS
  2018)}, Nov 2018.

\bibitem{ref:dataset}
H.~Aan{\ae}s, R.~R. Jensen, G.~Vogiatzis, E.~Tola, and A.~B. Dahl,
  ``Large-scale data for multiple-view stereopsis,'' \emph{International
  Journal of Computer Vision}, pp. 1--16, 2016.

\bibitem{ref:flow}
A.~{Dosovitskiy}, P.~{Fischer}, E.~{Ilg}, P.~{Häusser}, C.~{Hazirbas},
  V.~{Golkov}, P.~v.~d. {Smagt}, D.~{Cremers}, and T.~{Brox}, ``Flownet:
  Learning optical flow with convolutional networks,'' in \emph{2015 IEEE
  International Conference on Computer Vision (ICCV)}, Dec 2015, pp.
  2758--2766.

\bibitem{ref:photo}
C.~{Collewet} and E.~{Marchand}, ``Photometric visual servoing,'' \emph{IEEE
  Transactions on Robotics}, vol.~27, no.~4, pp. 828--834, Aug 2011.

\bibitem{ref:learnmove}
P.~{Agrawal}, J.~{Carreira}, and J.~{Malik}, ``Learning to see by moving,'' in
  \emph{2015 IEEE International Conference on Computer Vision (ICCV)}, Dec
  2015, pp. 37--45.

\bibitem{ref:imagenet}
J.~Deng, W.~Dong, R.~Socher, L.-J. Li, K.~Li, and L.~Fei-Fei, ``{ImageNet: A
  Large-Scale Hierarchical Image Database},'' in \emph{CVPR09}, 2009.

\bibitem{ref:goturn}
D.~Held, S.~Thrun, and S.~Savarese, ``Learning to track at 100 fps with deep
  regression networks,'' in \emph{European Conference on Computer Vision
  (ECCV)}.\hskip 1em plus 0.5em minus 0.4em\relax Springer, October 2017.

\bibitem{ref:regnet}
N.~{Schneider}, F.~{Piewak}, C.~{Stiller}, and U.~{Franke}, ``Regnet:
  Multimodal sensor registration using deep neural networks,'' in \emph{2017
  IEEE Intelligent Vehicles Symposium (IV)}, June 2017, pp. 1803--1810.

\end{thebibliography}
\end{document}